%% file: ms.tex
\documentclass{bmvc2k}
\usepackage{tikz}
\usepackage{standalone}
\usepackage{amssymb}

\title{Deep Stacked Networks with\\Residual Polishing for Image Inpainting}

\addauthor{Ugur Demir}{ugurdemir@itu.edu.tr}{1}
\addauthor{Gozde Unal}{unalgo@itu.edu.tr}{1}

\addinstitution{
 Department of Computer Engineering\\
 Istanbul Technical University\\
 Istanbul, Turkey
}

\runninghead{Demir, Unal}{Residual Polishing for Image Inpainting}

\def\etal{\emph{et al}\bmvaOneDot}

\begin{document}

\maketitle

\begin{abstract}
Deep neural networks have shown promising results in image inpainting even if the missing area is relatively large. However, most of the existing inpainting networks introduce undesired artifacts and noise to the repaired regions. To solve this problem, we present a novel framework which consists of two stacked convolutional neural networks that inpaint the image and remove the artifacts, respectively. The first network considers the global structure of the damaged image and coarsely fills the blank area. Then the second network modifies the repaired image to cancel the noise introduced by the first network. The proposed framework splits the problem into two distinct partitions that can be optimized separately, therefore it can be applied to any inpainting algorithm by changing the first network. Second stage in our framework which aims at polishing the inpainted images can be treated as a denoising problem where a wide range of algorithms can be employed. Our results demonstrate that the proposed framework achieves significant improvement on both visual and quantitative evaluations.
\end{abstract}

\section{Introduction}
\label{sec:intro}
The goal of inpainting is reconstruction of an image without incurring noticeable changes \cite{bertalmio_inpaint}. It is a widely used technique by the photo and video editing applications for repairing damaged images, removing undesired objects or refilling the missing parts of images. Although fixing the small deteriorations are relatively simple, filling the large holes or removing an object from the scene are still challenging due to complexity of the problem. 

With the recent advancement of Convolutional Neural Networks (CNN), several generative models that produce visually pleasant outputs have been presented for inpainting \cite{context_encoder, high_res_mc, ppgan}. The most popular approach is using an Autoencoder-like (AE) architecture that takes center cropped images (see Figure \ref{fig:samples}) and tries to synthesize realistic image patches to fill the blank areas. Results demonstrate that CNNs have a great potential to learn structure of the images collected from the real world \cite{alexnet, gan}.

One of the essential questions about realistic texture synthesis is: how can we measure the realism? No magical mathematical formula to determine whether an image is real or artificially constructed exists.  In order to solve this challenging problem, a crucial step is to construct synthesis models which are trained based on a comparison of real images with generated outputs. Although primitive objective functions like Euclidean Distance assist in measuring and comparing information on the general structure of the image, they tend to converge to the mean of pixel values that cause blurry outputs. 

Goodfellow \etal have taken image synthesis step forward by presenting Generative Adversarial Networks (GAN) \cite{gan}. An additional binary classifier, which is called a discriminative network, is included in order to classify whether an image comes from a real distribution or a generator network output. The discriminative network is trained  while the generative network tries to convince the former by producing more realistic patches. During the training, the generative network is scored by an adversarial loss that is calculated by the discriminator network. Another remarkable approach for image generation is given by a loss function that compares the features extracted from a pre-trained network instead of through direct pixel-wise measurements. It is known as the content loss in the literature \cite{perceptual,beyond_ae}. The idea of this approach lies in the recovery of high frequency details rather than blur. While using these loss functions, although plausible image patches are produced, they introduce undesired artifacts and noise due to complexity of the training procedure as can be seen in Figure \ref{fig:samples}.

The contribution of our work can be summarized upfront as follows: we divide the problem of image inpainting into two parts. First part, which we call the Coarse Painter Net (CPN), synthesizes the image texture by studying the uncorrupted parts. Second part of our method, which we call the Fine Painter Network (FPN) takes the reconstructed region and applies an enhancement to obtain an improved reconstruction. The latter is similar in spirit to the super-resolution problem, however, instead of enlarging the image, we aim to recover details without resizing. Furthermore, the second network aims to reduce noise if present. Overall, our FPN is designed to learn changes instead of the final image patch itself.

\begin{figure*}
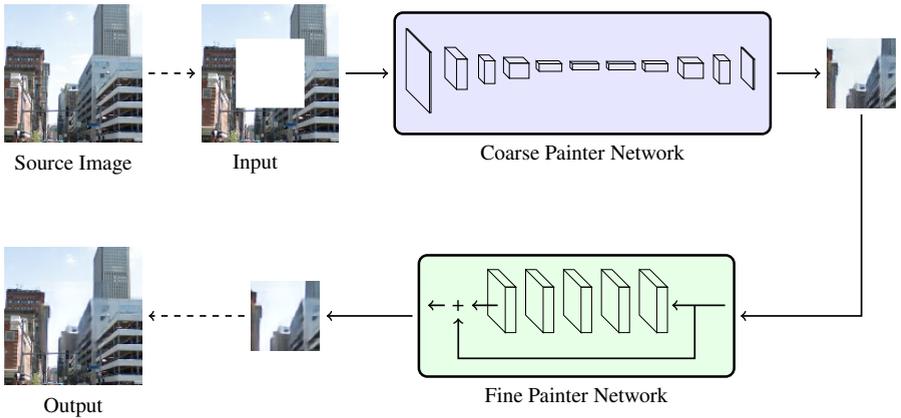

\begin{center}
\includestandalone[scale=0.8]{./images/gen_structure}
\end{center}
   \caption{General structure of the Residual Polishing framework.}
\label{fig:flow}
\end{figure*}


\section{Related Work}
As described above, we perform an inpainting process followed by an enhancement of the result by another process, overall of which we call the Residual Polishing framework. Thus, our work is related to a set of topics such as inpainting, denoising and super-resolution in the literature. 

Early studies on inpainting generally worked on a single damaged image \cite{bertalmio_inpaint, crimsi, graph_cut}. Starting from the border of the missing region, similar patches are searched inside the uncorrupted image segments. The closest discovered patches are placed into the missing region, and the process is repeated until the blank area is completely filled. Most of those earlier works paid attention to improving the speed of similar patch finding procedure. Although those methods have shown superior performance on the images which contain similar texture details, they suffered from the lack of global structural information, which led to undesirable outputs.

CNN has shown great success in both classification \cite{vgg, inception, resnet} and regression tasks \cite{segmentation, gan, dcgan}. Especially AE architecture has been used widely for image generation and reconstruction \cite{context_encoder, srcnn, srgan}. Adversarial training scheme has shown striking success for realistic image generation. Lately, a wide range of GAN type architectures, which offer novel objective functions by changing the structure of discriminator \cite{ebgan, ppgan, categor_gan}, have been proposed. Pathak \etal use the GAN for an inpainting CNN which is called Context-Encoder \cite{context_encoder}. Their inpainting network takes the center cropped images and regress the missing part. Indeed, our CPN architecture is inspired from the Context-Encoder. On the other hand, most GANs have been focusing solely on the realistic image generation instead of generation of an image patch well-matched to the global image, and that property of GANs is incompatible with the original goal of the inpainting.

Using features extracted from pre-trained networks for comparing images has become popular recently in art transfer \cite{perceptual, mcmc}, AE training \cite{beyond_ae} and inpainting \cite{high_res_mc}. In the literature, this idea is known as either mainly content loss, perceptual loss, style loss or feature loss, as we call here. Generally, features are extracted from a classification network, which is trained on huge data collections due to their representation strength. During the optimization, features of the generated image and the ground truth are forced to be close. The learned feature loss measures produce more robust results compared to those of the pixel-based distance measures \cite{beyond_ae}. Notable inpainting results are obtained recently through using a combination of the feature loss and the adversarial loss \cite{semantic_inpaint}. Also, in \cite{high_res_mc}, Yang \etal reported state-of-the-art results by applying a local texture constraint that was inspired from \cite{mcmc} along with the content loss and the adversarial loss.

Residual Networks have been developed to improve gradient flow in the deep networks through addition of skip connections to the architecture \cite{resnet, identity_map, wide_res, inception, aggregated_res}. They stack the residual blocks which consist of several convolutional layers and a residual connection. This operation reduces the training time dramatically while classification accuracies on the public datasets are improved significantly. In another area of regression problems, residual connections are used differently. Kim \etal proposed a super-resolution model that learns the difference between the high resolution image and its low resolution counterparts \cite{vdsr}. The final result is obtained by adding the difference to the input image. It was shown that regressing the difference between input and the desired output can produce considerable performance improvement \cite{dncnn, ct_reconstruct, darn}. 

In this paper, we present a novel method that combines a coarse painter and a fine painter network, which is described in Section 3. Our fine painter idea is inspired from the super-resolution problem where a high resolution image is produced based on a given low resolution input. To our knowledge, our proposed residual polishing method is the first application that uses the residual connections to improve the results in image inpainting problem, which we will demonstrate in Experiments (Section \ref{sec:exp}), followed by Conclusions (Section \ref{sec:conc}).


\section{Residual Polishing}
In Residual Polishing framework, the intention is completion of the inpainting process at two stages. In the beginning, we obtain the input image $\tilde{x}$ by removing the center part of the image $x$ that is taken from the dataset. Our CPN fills the blank regions in $\tilde{x}$ to obtain inpainted image part $\tilde{y}$ at the end of the first stage. We hypothesize that $\tilde{y}$ has noise and artifacts introduced by the CPN, therefore, further improvement should be possible by applying a suitable procedure. To improve quality of the ultimate result, in the second stage, the FPN removes the undesired effects on $\tilde{y}$ and generates the final image $y$. Figure \ref{fig:flow} shows the general structure of our approach. Apart from noise and artifacts, FPN additionally finds the undiscovered details. The generated image part $y$ is placed to the missing area of the input $x$ to finalize the algorithm. Here we can formulate Residual Polishing as;
\begin{equation}
y = F(C(M(x))),
\end{equation}
where $F(\cdot)$ represents FPN, $C(\cdot)$ represents CPN and $M(\cdot)$ is the center removal operation to obtain $\tilde{x}$ from $x$. Following sections give details about the architectures and training steps.

\subsection{Coarse Painter Network}
CPN is formed by sequentially stacking an encoder and a decoder module . The Encoder part takes an input image $\tilde{x}$ and produces a latent representation called the bottleneck features. The latent output is passed to the decoder network to generate missing part of $\tilde{x}$. This approach is inspired from the Context-Encoder proposed in \cite{context_encoder}. The difference is that we use a fully-connected layer at the end of the encoder instead of the channel-wise fully-connected layer.

\begin{figure}[h]
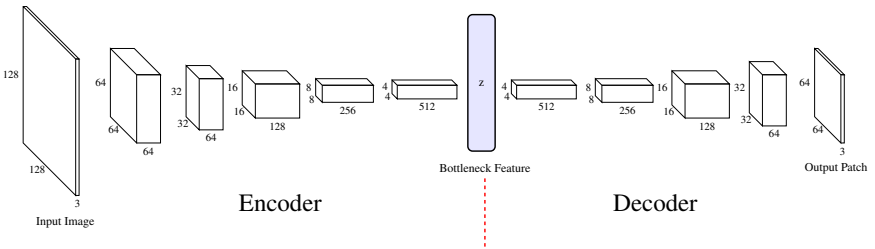

\begin{center}
\includestandalone[scale=0.45]{./images/coarse_arch}
\end{center}
   \caption{Coarse Painter Network architecture.}
\label{fig:coarse_arch}
\end{figure}

Architecture of the CPN is very similar to that of \cite{dcgan}. The filter sizes are fixed to 4x4 for each convolutional layer. To expose global information in the image, input is subsampled by strided convolutions. Filter depth is doubled for each subsampling operations. Layers of the encoder other than the last one consist of convolution, batch normalization \cite{batch_norm} and Leaky ReLU (LReLU) activation, respectively. The last layer contains only a traditional fully-connected layer which connects the encoder to the decoder. To reconstruct the output from the bottleneck features, decoder applies transposed convolution, batch normalization and Exponential Linear Unit (ELU) \cite{elu} activation. Figure \ref{fig:coarse_arch} shows the detailed architectural design of the CPN.

At the training stage, we use a combination of three loss functions. They are optimized jointly via backpropagation. We describe each loss function briefly as follows.

{\bf Euclidean Loss} computes the pixel-wise Euclidean Distance between the synthesized image patch and the ground truth. Even though it forces the network to produce a blurry output, it guides the network to roughly predict texture colors. It is defined as:
\begin{equation}
\mathcal{L}_{ed} = \frac{1}{N}\sum_{n=1}^N\frac{1}{WHC}\sqrt{(\tilde{y}_n-x_n^{center})^2}
\end{equation}
where $N$ is the number of samples, $x_n^{center}$ is the ground truth, $\tilde{y}$ is the generated output, $W$, $H$, $C$ are width, height and channel of the compared images. 

{\bf Adversarial Loss} is computed by the discriminator network $\mathcal{D}$ that is introduced in the training phase. It tries to distinguish whether the input comes from the real data distribution $p_{train}(x^{center})$  or generator output distribution $p_{C}(\tilde{x})$. The generative network $\mathcal{C}$, which is CPN in that case, is optimized to fool the $\mathcal{D}$ while the discriminator tries to increase its accuracy. $\mathcal{C}$ and $\mathcal{D}$ are trained simultaneously by solving
\begin{equation}
\label{eq:adv_minmax}
\operatorname*{min}_{\theta_C} \operatorname*{max}_{\theta_D} V(C,\mathcal{D}) = \mathbb{E}_{x^{center}\sim p_{train}(x^{center})}[\log \mathcal{D}(x^{center})] + \mathbb{E}_{\tilde{y}\sim p_{C}(\tilde{x})}[\log (1 - \mathcal{D}(C(\tilde{x})))]
\end{equation}
where $\theta_C$ and $\theta_D$ are the parameters of the CPN and the discriminator network. While the adversarial loss helps to generate more realistic image textures, it causes appearance of superfluous details. Relying only on the adversarial loss makes training difficult and causes unstable behaviour.

{\bf Feature Loss} transfers the compared images from the pixel value space to a space where the features obtained from an external model are used. We utilized VGG16 \cite{vgg} network, which is trained on ImageNet dataset \cite{imagenet}, as the feature extraction network due to its proven success. The feature network is used with pre-trained values and its weights are kept constant during the training. We use the intermediate activation maps (relu\_2\_1) as features. The feature loss is calculated by
\begin{equation}
\label{eq:feat_loss}
\mathcal{L}_{feat} = \frac{1}{N}\sum_{n=1}^N\frac{1}{WHC}\sqrt{(\Phi(y_n')-\Phi(x_n))^2}
\end{equation}
where $y_n'$ is the inpainted image, $\Phi(\cdot)$ is the feature extraction operation, $W$, $H$, $C$ are the width, height and depth of the activation map. Here, we note that in Equation \ref{eq:feat_loss}, the whole inpainted image $y_n'$ and the whole source image $x_n$ are compared instead of just the center part of the original image and the output of the CPN to make use of the global structure similarity in images.

The final CPN architecture and the training strategy are determined through the experiments. The best results are obtained when the combination of the Euclidean Loss, Adversarial Loss and Feature Loss are used as the objective function. Each component of the loss function is governed by a coefficient $\lambda$:
\begin{equation}
\label{eq:loss_cpn}
\mathcal{L}_C = \lambda_{ed} \mathcal{L}_{ed} + \lambda_{adv} \mathcal{L}_{adv} + \lambda_{feat} \mathcal{L}_{feat}
\end{equation}
Adversarial loss $\mathcal{L}_{adv}$ is calculated by solving the Equation \ref{eq:adv_minmax}. Also a L2 regularization term is added to the loss function to apply weight decay to the CPN parameters.

\subsection{Fine Painter Network}
\begin{figure}[h]
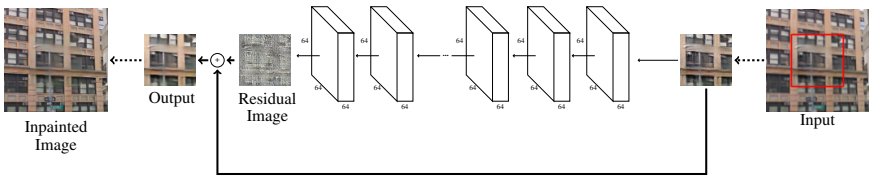

\begin{center}
\includestandalone[scale=0.3]{./images/fine_arch}
\end{center}
   \caption{Fine Painter Network architecture.}
\label{fig:fine_arch}
\end{figure}

Our FPN in Figure \ref{fig:fine_arch} takes the image patch obtained from the CPN, and supposing that the output of the CPN has a noisy characteristic, it aims to improve its quality by a "noise-removal" operation through residual connections in the network.

Input and output of the FPN are close to each other because most of the texture detail is determined by the first network. Thus, learning the residual image which is the difference between the input and the output is more accessible than directly regressing the output. To condition the network to produce a residual image $r$, the input of the FPN is connected to the output with a skip connection. Defining the residual image by $r=x^{center}-\tilde{y}$, the objective function becomes 
\begin{equation}
\label{eq:loss_fine}
\mathcal{L}_F = \frac{1}{N}\sum_{n=1}^N\frac{1}{WHC}\sqrt{(r_n-x_n^{center})^2}.
\end{equation}

Equation \ref{eq:loss_fine} indicates that residual image $r$ and the ground truth $x^{center}$ must have the same size. In order to satisfy the size constraint, the convolutional layers are used without stride and activation maps are padded with zero. We build two different FPNs. First version uses only cascaded convolutional layers and ELU activations. Our second design puts batch normalization between convolution and the activation for each layer. The second network was experimented with several activation functions and we obtained the best results with ReLU. Figure \ref{fig:fine_arch} shows our FPN architecture.

To obtain the polished output, the generated residual image is added to the output of the CPN. Similar to the literal meaning of "polishing", our Residual Polishing network grinds the surface of the image to make it appear smoother with reduced artifacts in the inpainted area.


\section{Experiments}
\label{sec:exp}
In this section, we evaluate the performance of our proposed method and compare residual polishing with the recent inpainting methods. One of the major problems of inpainting applications is measuring the output quality. Whereas the extracted center parts of the images are used as ground truths, the generated patches can be different while they are still plausible. Thus, pixel-wise comparison can be misleading. Nevertheless, to compare the algorithms, we use peak signal to noise ratio (PSNR), mean L1 and L2 losses for evaluation. We also provide visual evidence of successfully inpainted images in Figure \ref{fig:samples}.

\subsection{Dataset}
With the advancement of augmented and virtual reality applications, street view images and videos receive increased interest. People can travel around the world even without being there. However, a large amount of confidential or private scenes exists in the street view image collections. To avoid personal privacy breach, for instance, Google\textsuperscript{TM} adds blur or some filtering to cover undesired image parts, which certainly disrupts the integrity of user experience. As per mentioned motivation, we trained our proposed inpainting network on Google Street View dataset \cite{crcv} to learn realistic street view image generation so that we can hide the unwanted parts in the images.

Google Street View dataset consist of 62058 high quality images. It is divided into 10 parts. We use the first and tenth parts as the testing set, the ninth part for validation, and the rest of the parts are included in the training set. In this way, 46200 images are used for training. Images are scaled to the size 128x128 and its 64x64 sized center part is cropped and kept as the ground truth. We do not apply data augmentation at the training stage.

\subsection{Implementation}
Residual Polishing framework is implemented using Tensorflow \cite{tensorflow}. Our networks are trained separately on NVIDIA\textsuperscript{TM} Tesla K20 5GB and GeForce\textsuperscript{TM} GTX 960 4GB graphic cards. For performance measurements, we used Context-Encoder Torch \cite{torch}  implementation provided by its authors. Authors of \cite{high_res_mc} have made available only the pre-trained model of their approach. Therefore, we could not train that network on the Google Street View dataset. Instead, to our CPN output, we applied a local texture constraint that is used by the mentioned method, and obtained plausible results.

\begin{figure}[h]
\begin{center}
\begin{tabular}{cc}
\bmvaHangBox{\includegraphics[width=6.1cm]{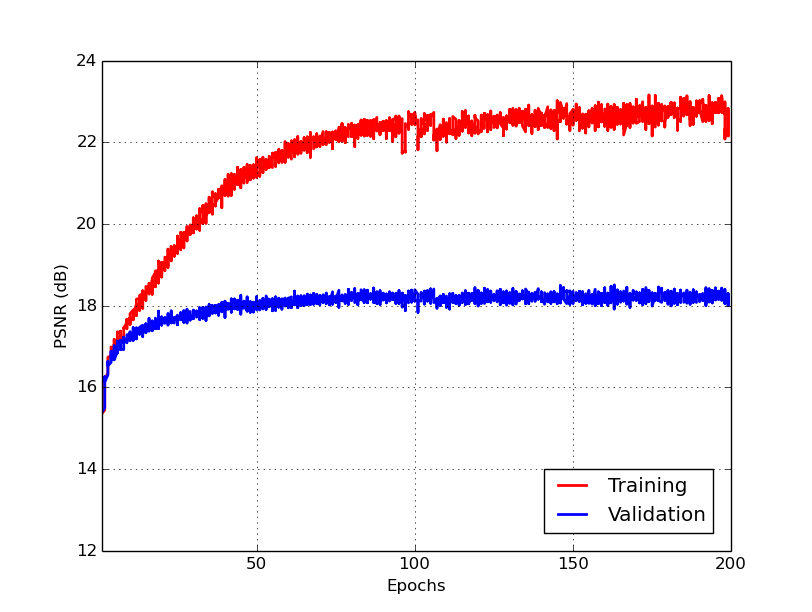}}
&
\bmvaHangBox{\includegraphics[width=6.1cm]{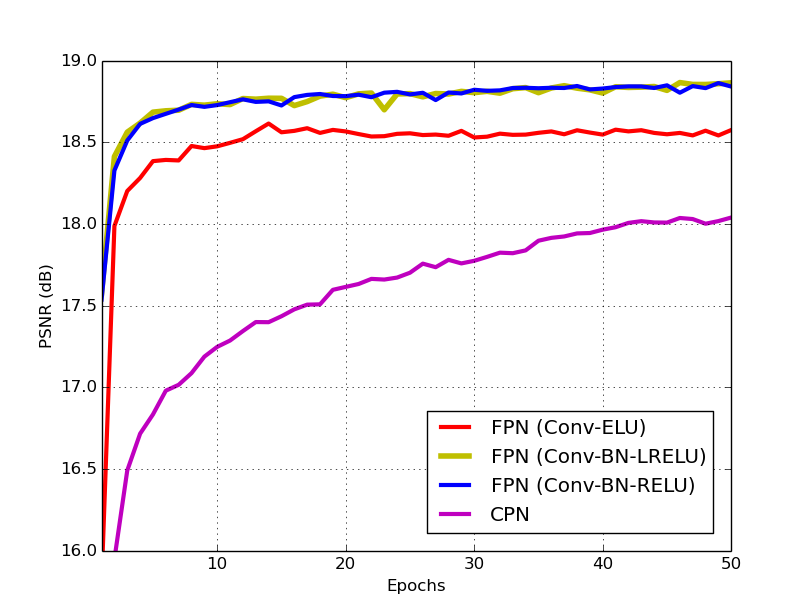}}\\
(a) Coarse Painter Network & (b) Fine Painter Networks on Validation Set
\end{tabular}
\end{center}
\caption{PSNR curve of the networks: (a) CPN performance on training and validation set; (b) Comparison of different FPN architectures and CPN on evaluation set. Note that CPN is not trained during the FPN training, we added CPN curve on (b) for comparison purposes.}
\label{fig:curves}
\end{figure}

\subsection{Training}
CPN is trained with joint loss function stated in Equation \ref{eq:loss_cpn} using Adam optimizer \cite{adam}. We set the parameters for the optimizer as $\beta_1=0.5$, $\beta_2=0.999$ and $\epsilon=10^{-8}$. Contributions of different loss functions are determined by the parameters $\lambda_{ed}=0.5$, $\lambda_{ed}=0.001$ and $\lambda_{feat}=0.0001$. Figure \ref{fig:curves} shows the training and validation performance of CPN during 200 epochs. It is clearly seen that after 50 epochs, CPN stops learning. Thus, we take the model at that point as our final CPN.

During the FPN training, CPN weights are not updated. FPN is trained by Adam optimizer which uses the same parameters specified for CPN. In Figure \ref{fig:curves}, we show the performance of different FPN architectures against CPN. We tried several residual architectures with different activations. Without batch normalization, ELU is the only activation that we can train our network where ReLU and LReLU could not achieve considerable improvements. After batch normalization layers are placed between convolution layers and the activations, we obtain the best results for our setup. With batch normalization, performance of ReLU and LReLU are too close. The final FPN is constructed by convolution, batch normalization and ReLU blocks.

\begin{figure}[t]
\begin{center}
\begin{tabular}{c@{\hskip 0.1cm}c@{\hskip 0.1cm}c@{\hskip 0.1cm}c@{\hskip 0.1cm}c@{\hskip 0.1cm}}
(a) Input & (b) Ctx-Enc & (c) NPS & (d) CPN & (e) Our Results \\
\includegraphics[width=2.1cm]{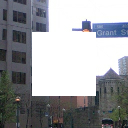} &
\includegraphics[width=2.1cm]{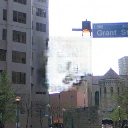}&
\includegraphics[width=2.1cm]{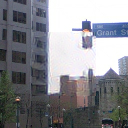}&
\includegraphics[width=2.1cm]{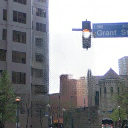}&
\includegraphics[width=2.1cm]{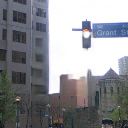} \\

\includegraphics[width=2.1cm]{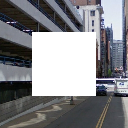}&
\includegraphics[width=2.1cm]{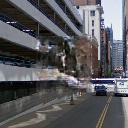}&
\includegraphics[width=2.1cm]{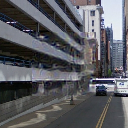}&
\includegraphics[width=2.1cm]{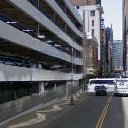}&
\includegraphics[width=2.1cm]{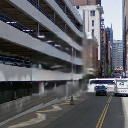} \\

\includegraphics[width=2.1cm]{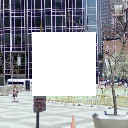}&
\includegraphics[width=2.1cm]{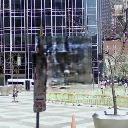}&
\includegraphics[width=2.1cm]{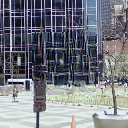}&
\includegraphics[width=2.1cm]{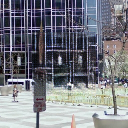}&
\includegraphics[width=2.1cm]{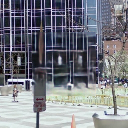} \\

\includegraphics[width=2.1cm]{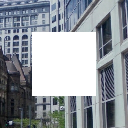}&
\includegraphics[width=2.1cm]{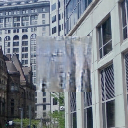}&
\includegraphics[width=2.1cm]{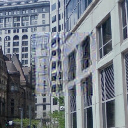}&
\includegraphics[width=2.1cm]{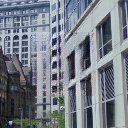}&
\includegraphics[width=2.1cm]{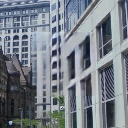}\\

\end{tabular}
\end{center}
\caption{Image inpaintig results obtained from different methods. Please zoom in while comparing the results. (a) Center cropped input images taken from the test set; (b) Context Encoder \cite{context_encoder} output; (c) Output of Neural Patch Synthesis \cite{high_res_mc}; (d) CPN output (e) FPN output.}
\label{fig:samples}
\end{figure}

\begin{figure}[t]
\begin{center}
\begin{tabular}
{c@{\hskip 0.01cm}c@{\hskip 0.01cm}c@{\hskip 0.01cm}c@{\hskip 0.01cm}
c@{\hskip 0.01cm}c@{\hskip 0.01cm}c@{\hskip 0.01cm}c@{\hskip 0.01cm}}

\includegraphics[scale=0.32]{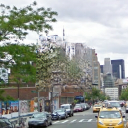}&
\includegraphics[scale=0.32]{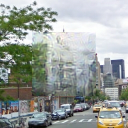}&
\includegraphics[scale=0.32]{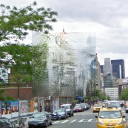}&
\includegraphics[scale=0.32]{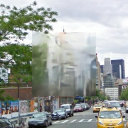}&

\includegraphics[scale=0.32]{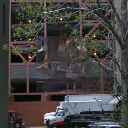}&
\includegraphics[scale=0.32]{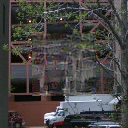}&
\includegraphics[scale=0.32]{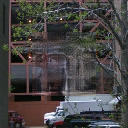}&
\includegraphics[scale=0.32]{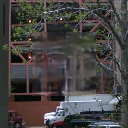}\\

\includegraphics[scale=0.32]{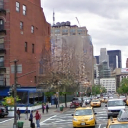}&
\includegraphics[scale=0.32]{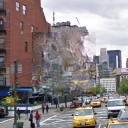}&
\includegraphics[scale=0.32]{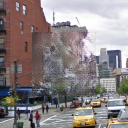}&
\includegraphics[scale=0.32]{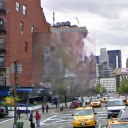}&

\includegraphics[scale=0.32]{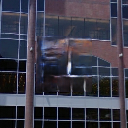}&
\includegraphics[scale=0.32]{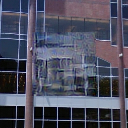}&
\includegraphics[scale=0.32]{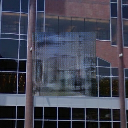}&
\includegraphics[scale=0.32]{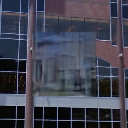}\\

(a)&(b)&(c)&(d)&(e)&(f)&(g)&(h)\\
\end{tabular}
\end{center}
\caption{Fail cases of Context Encoder \cite{context_encoder} (a,e), Neural Patch Synthesis \cite{high_res_mc}  (b,f), CPN (c,g) and FPN (d,h).}
\label{fig:fails}
\end{figure}

\subsection{Evaluations}
We evaluate the performance of our algorithm against recent inpainting algorithms \cite{high_res_mc, context_encoder}. The output images are given in Figure \ref{fig:samples}. Visual outputs show that our Residual Polishing approach softens the inpainted image and produces visually plausible outputs. In Table \ref{aa}, we demonstrate that our method achieves the best PSNR value on Google Street View dataset. 

\begin{table}[h]
\begin{center}
\begin{tabular}{|l|c|c|c|}
\hline
Method & Mean L1 Loss & Mean L2 Loss & PSNR \\
\hline\hline
Context-Encoder \cite{context_encoder} & 2.74 & 0.53 & 20.60 dB \\
\hline
Neural Patch Synth.\cite{high_res_mc} & 5.74 & 1.01 & 20.72 dB \\
\hline
CPN (Our) & 1.97 & 0.38 & 21.37 dB\\
{\bf Residual Polish (Our)} & {\bf 1.74} & {\bf 0.32} & {\bf 22.89 dB}\\
\hline
\end{tabular}
\end{center}
\caption{Performance comparison on Google Street View dataset. For each measures the best results are shown in bold.}
\label{aa}
\end{table}

Our results show that instead of proposing an end-to-end solution, dividing the inpainting problem into simpler tasks and solving them separately can produce better results. Although our CPN architecture is inspired from Context-Encoder and it has nearly the same number of parameters, CPN achieves better results due to addition of the feature loss. This indication is consistent with the recent studies \cite{beyond_ae, perceptual, mcmc, semantic_inpaint} and shows that using pre-trained network features improves texture synthesis quality.

Residual polishing policy provides us additional performance gain by fixing the local deformations introduced by previous inpainting network. If the first network does not generate a proper texture, our FPN cannot improve the texture details (see Figure \ref{fig:fails} for example fail cases). On the other hand, if we increase the receptive field of our residual network, it can be capable of repairing more global deformations as stated in \cite{vdsr}.

\section{Conclusion}
\label{sec:conc}
In this paper, we proposed Residual Polishing framework as a novel inpainting algorithm. Our motivation is to simplify the inpainting problem by dividing it into two stages which for each we present different networks. At the first stage, a coarse texture is obtained by considering at the surrounding pixels of the damaged area. Then our second network produces a residual image which is the difference between the desired output and the coarsely inpainted image. We have demonstrated that this residual image contains significant information that improves the performance of our final results. Residual Polishing framework can be benefited by any of the inpainting algorithms to enhance their outputs. Further, we will investigate different architectures and training policies for our residual network to ensure it can fix even more complex artifacts.
  

\bibliography{egbib}
\end{document}

%% file: images/gen_structure.tex
\newcommand{\layerold}[1]{
	\coordinate (start) at #1;
	\def \x {0.4}
	\def \y {4}
	\def \z {4}

	\draw[black] (start) -- ++(-\x,0,0) -- ++(0,-\y,0) -- ++(\x,0,0) -- cycle;
	\draw[blue] (start) -- ++(0,0,-\z) -- ++(0,-\y,0) -- ++(0,0,\z) -- cycle;
	\draw[green] (start) -- ++(-\x,0,0) -- ++(0,0,-\z) -- ++(\x,0,0) -- cycle;
}

\newcommand{\cc}[1]{
	\coordinate (start) at #1;
	\def \x {0.4}
	\def \y {2}
	\def \z {1.25}

	\draw[black] (start) -- ++(-\x,0,0) -- ++(0,-\y,0) -- ++(\x,0,0) -- cycle;
	\draw[blue] (start) -- ++(0,0,-\z)  -- ++(0,-\y,0) node[right, pos=0.5]{\small 256} -- ++(0,0,\z) -- cycle;
}

\newcommand{\bb}[1]{
	\coordinate (start) at #1;

	\def \x {4}
	\def \y {1.8}
	\def \z {2.5}

	\draw [draw=blue, every edge/.append style={draw=blue, densely dashed, opacity=.5}, fill=cyan]
    (start) coordinate (o) -- ++(-\x,0,0) coordinate (a) -- ++(0,-\y,0) coordinate (b) 
    edge coordinate [pos=1] (g) ++(0,0,-\z) node[pos=0.5] {sd} -- ++(\x,0,0) coordinate (c) -- cycle
    (o) -- ++(0,0,-\z) coordinate (d) -- ++(0,-\y,0) coordinate (e) edge (g) -- (c) -- cycle
    (o) -- (a) -- ++(0,0,-\z) coordinate (f) edge (g) -- (d) -- cycle;


}

\newcommand{\layer}[7]{
	\coordinate (start) at #1;
	\def \hs {#2}
	\def \ws {#3}
	\def \cs {#4}
	
	\begin{scope}[rotate around={90:(start)}]
		\draw[black] (start) -- ++(0,-\cs,0) -- ++(0,0,-\ws) -- ++(0,\cs,0);
		\draw[black] (start) -- ++(0,0,-\ws) -- ++(-\hs,0,0) node[left, pos=0.5]{\small #5} -- ++(0,0,\ws) node[left, pos=0.5]{\small #6} -- cycle;
		\draw[black] (start) -- ++(0,-\cs,0) -- ++(-\hs,0,0) -- ++(0,\cs,0) node[below, pos=0.5]{\small #7};
	\end{scope}
}

\newcommand{\la}[4]{
	\coordinate (start) at #1;
	\def \hs {#2}
	\def \ws {#3}
	\def \cs {#4}
	
	\begin{scope}[rotate around={90:(start)}]
		\draw[black] (start) -- ++(0,-\cs,0) -- ++(0,0,-\ws) -- ++(0,\cs,0);
		\draw[black] (start) -- ++(0,0,-\ws) -- ++(-\hs,0,0) -- ++(0,0,\ws) -- cycle;
		\draw[black] (start) -- ++(0,-\cs,0) -- ++(-\hs,0,0) -- ++(0,\cs,0);
	\end{scope}
}

\newcommand{\cnet}{

	\coordinate (L1) at (-0.5, 0, 0);
	\coordinate (L2) at (1.75, 0, 1.2);
	\coordinate (L3) at (3.7 , 0, 1.5);
	\coordinate (L4) at (5.5 , 0, 1.9);
	\coordinate (L5) at (7.5 , 0, 2);
	\coordinate (L6) at (9.7 , 0, 2);
	
	\coordinate (L7) at (12.1, 0, 2);
	\coordinate (L8) at (14.5, 0, 2);
	\coordinate (L9) at (17.0, 0, 1.9);
	\coordinate (L10) at (19.2, 0, 1.5);
	\coordinate (L11) at (21.3, 0, 1.2);

    \la{(L1)}{4}{4}{0.1}
    \la{(L2)}{2}{2}{0.7}
    \la{(L3)}{1.5}{1}{0.7}
    \la{(L4)}{1}{1}{1.3}
    \la{(L5)}{0.5}{0.5}{1.5}
    \la{(L6)}{0.4}{0.4}{1.75}

    \la{(L7)}{0.4}{0.4}{1.75}
    \la{(L8)}{0.5}{0.5}{1.5}
    \la{(L9)}{1}{1}{1.3}{16}
    \la{(L10)}{1.5}{1}{0.7}
    \la{(L11)}{2}{2}{0.1}
}

\newcommand{\fnet} {
	\coordinate (F1) at (1.5, 0, 0);
	\coordinate (F2) at (4.0, 0, 0);
	\coordinate (F3) at (6.5 ,0, 0);
	\coordinate (F4) at (9.0 ,0, 0);
	\coordinate (F5) at (11.5 ,0, 0);
	
	\la{(F1)}{3}{3}{0.7}
	\la{(F2)}{3}{3}{0.7}
	\la{(F3)}{3}{3}{0.7}
	\la{(F4)}{3}{3}{0.7}
	\la{(F5)}{3}{3}{0.7}

	\path (-1.6,-1.2) node[](y) {+};
	\draw[->, black, thick] (16,-1.2,0) -- ++(-3.5,0,0);
	\draw[->, black, thick] (1,-1.2,0) -- (y);
	\draw[->, black, thick] (14,-1.2,0) -- ++(0,-3.5,0) -- ++(-15.6,0,0) -- (y);
	\draw[->, black, thick] (y) -- ++(-2,0,0);
	
}

\begin{tikzpicture}








\coordinate (C1) at (2.3, -1);
\def \ww {6.2}
\def \hh {2.0}

\node[inner sep=3pt] (inp) at (-3,0) 
	{\includegraphics[scale=0.5]{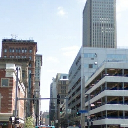}};

\node[inner sep=3pt] (ctx) at (0.25,0) 
	{\includegraphics[scale=0.5]{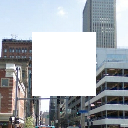}};

\node[inner sep=3pt] (cpn_center) at (10,0) 
	{\includegraphics[scale=0.5]{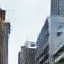}};

\node[inner sep=3pt] (fpn_c) at (0.5,-4) 
	{\includegraphics[scale=0.5]{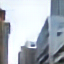}};

\node[inner sep=3pt] (fpn_f) at (-3,-4) 
	{\includegraphics[scale=0.5]{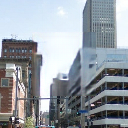}};

\node[rectangle, rounded corners, draw=black, very thick, fill=blue!10, minimum width=6.2cm, minimum height=2cm, text centered, outer sep=3pt] (cpn) at (5.4,0) 
		{};

\node[rectangle, rounded corners, draw=black, very thick, fill=green!10, minimum width=5.2cm, minimum height=2cm, text centered, outer sep=3pt] (fpn) at (5.3,-4) 
		{};

\node[] at (-3,-1.5) {Source Image};
\node[] at (0,-1.5) {Input};
\node[] at (5.4,-1.3) {Coarse Painter Network};
\node[] at (5.3, -5.3) {Fine Painter Network};
\node[] at (-3,-5.5) {Output};

\draw[->,thick, dashed] (inp) -- (ctx);
\draw[->,thick] (ctx) -- (cpn);
\draw[->,thick] (cpn) -- (cpn_center);
\draw[->,thick] (cpn_center) |- (fpn.east);
\draw[->,thick] (fpn) -- (fpn_c);
\draw[->,thick, dashed] (fpn_c) -- (fpn_f);

\begin{scope}[scale=0.25, shift={(12,1.4,0)}]
\cnet{(10,10,10)}
\end{scope}

\begin{scope}[scale=0.25, shift={(15,-14.1,0)}]
\fnet
\end{scope}

\end{tikzpicture}

%% file: images/coarse_arch.tex
\newcommand{\layerold}[1]{
	\coordinate (start) at #1;
	\def \x {0.4}
	\def \y {4}
	\def \z {4}

	\draw[black] (start) -- ++(-\x,0,0) -- ++(0,-\y,0) -- ++(\x,0,0) -- cycle;
	\draw[blue] (start) -- ++(0,0,-\z) -- ++(0,-\y,0) -- ++(0,0,\z) -- cycle;
	\draw[green] (start) -- ++(-\x,0,0) -- ++(0,0,-\z) -- ++(\x,0,0) -- cycle;
}

\newcommand{\cc}[1]{
	\coordinate (start) at #1;
	\def \x {0.4}
	\def \y {2}
	\def \z {1.25}

	\draw[black] (start) -- ++(-\x,0,0) -- ++(0,-\y,0) -- ++(\x,0,0) -- cycle;
	\draw[blue] (start) -- ++(0,0,-\z)  -- ++(0,-\y,0) node[right, pos=0.5]{\small 256} -- ++(0,0,\z) -- cycle;
}

\newcommand{\bb}[1]{
	\coordinate (start) at #1;

	\def \x {4}
	\def \y {1.8}
	\def \z {2.5}

	\draw [draw=blue, every edge/.append style={draw=blue, densely dashed, opacity=.5}, fill=cyan]
    (start) coordinate (o) -- ++(-\x,0,0) coordinate (a) -- ++(0,-\y,0) coordinate (b) 
    edge coordinate [pos=1] (g) ++(0,0,-\z) node[pos=0.5] {sd} -- ++(\x,0,0) coordinate (c) -- cycle
    (o) -- ++(0,0,-\z) coordinate (d) -- ++(0,-\y,0) coordinate (e) edge (g) -- (c) -- cycle
    (o) -- (a) -- ++(0,0,-\z) coordinate (f) edge (g) -- (d) -- cycle;


}

\newcommand{\layer}[7]{
	\coordinate (start) at #1;
	\def \hs {#2}
	\def \ws {#3}
	\def \cs {#4}
	
	\begin{scope}[rotate around={90:(start)}]
		\draw[black] (start) -- ++(0,-\cs,0) -- ++(0,0,-\ws) -- ++(0,\cs,0);
		\draw[black] (start) -- ++(0,0,-\ws) -- ++(-\hs,0,0) node[left, pos=0.5]{\small #5} -- ++(0,0,\ws) node[left, pos=0.5]{\small #6} -- cycle;
		\draw[black] (start) -- ++(0,-\cs,0) -- ++(-\hs,0,0) -- ++(0,\cs,0) node[below, pos=0.5]{\small #7};
	\end{scope}
}

\newcommand{\conv}[7]{
	
}

\begin{tikzpicture}
	\coordinate (L1) at (-0.5, 0, 0);
	\coordinate (L2) at (1.75, 0, 1.2);
	\coordinate (L3) at (3.7 , 0, 1.5);
	\coordinate (L4) at (5.5 , 0, 1.9);
	\coordinate (L5) at (7.5 , 0, 2);
	\coordinate (L6) at (9.7 , 0, 2);
	
	\coordinate (L7) at (13.2, 0, 2);
	\coordinate (L8) at (15.7, 0, 2);
	\coordinate (L9) at (18.1, 0, 1.9);
	\coordinate (L10) at (20.1, 0, 1.2);
	\coordinate (L11) at (22.4, 0, 1.2);

	\node[rectangle, rounded corners, draw=black, very thick, fill=blue!10, minimum width=0.8cm, minimum height=4cm, text centered] (disc) at (11.4,-0.7) 
		{z};


    \layer{(L1)}{4}{4}{0.1}{128}{128}{3}
    \layer{(L2)}{2}{2}{0.7}{64}{64}{64}
    \layer{(L3)}{1.5}{1}{0.7}{32}{32}{64}
    \layer{(L4)}{1}{1}{1.3}{16}{16}{128}
    \layer{(L5)}{0.5}{0.5}{1.5}{8}{8}{256}
    \layer{(L6)}{0.4}{0.4}{1.75}{4}{4}{512}

    \layer{(L7)}{0.4}{0.4}{1.75}{4}{4}{512}
    \layer{(L8)}{0.5}{0.5}{1.5}{8}{8}{256}
    \layer{(L9)}{1}{1}{1.3}{16}{16}{128}
    \layer{(L10)}{1.5}{1}{0.7}{32}{32}{64}
    \layer{(L11)}{2}{2}{0.1}{64}{64}{3}

    \node[text centered] (input) at (-0.8, -4.8) {Input Image};
    \node[text centered] (output) at (21.8, -3.2) {Output Patch};
    \node[text centered] (btl) at (11.5, -3.2) {Bottleneck Feature};

    \draw[-,dashed, red, very thick] (11.5, -3.5) -- ++(0,-2);

    \node[text centered] (btl) at (5.5, -4.2) {\huge Encoder};
    \node[text centered] (btl) at (16.5, -4.2) {\huge Decoder};



\end{tikzpicture}

%% file: images/fine_arch.tex
\newcommand{\layerold}[1]{
	\coordinate (start) at #1;
	\def \x {0.4}
	\def \y {4}
	\def \z {4}

	\draw[black] (start) -- ++(-\x,0,0) -- ++(0,-\y,0) -- ++(\x,0,0) -- cycle;
	\draw[blue] (start) -- ++(0,0,-\z) -- ++(0,-\y,0) -- ++(0,0,\z) -- cycle;
	\draw[green] (start) -- ++(-\x,0,0) -- ++(0,0,-\z) -- ++(\x,0,0) -- cycle;
}

\newcommand{\cc}[1]{
	\coordinate (start) at #1;
	\def \x {0.4}
	\def \y {2}
	\def \z {1.25}

	\draw[black] (start) -- ++(-\x,0,0) -- ++(0,-\y,0) -- ++(\x,0,0) -- cycle;
	\draw[blue] (start) -- ++(0,0,-\z)  -- ++(0,-\y,0) node[right, pos=0.5]{\small 256} -- ++(0,0,\z) -- cycle;
}

\newcommand{\bb}[1]{
	\coordinate (start) at #1;

	\def \x {4}
	\def \y {1.8}
	\def \z {2.5}

	\draw [draw=blue, every edge/.append style={draw=blue, densely dashed, opacity=.5}, fill=cyan]
    (start) coordinate (o) -- ++(-\x,0,0) coordinate (a) -- ++(0,-\y,0) coordinate (b) 
    edge coordinate [pos=1] (g) ++(0,0,-\z) node[pos=0.5] {sd} -- ++(\x,0,0) coordinate (c) -- cycle
    (o) -- ++(0,0,-\z) coordinate (d) -- ++(0,-\y,0) coordinate (e) edge (g) -- (c) -- cycle
    (o) -- (a) -- ++(0,0,-\z) coordinate (f) edge (g) -- (d) -- cycle;


}

\newcommand{\layer}[7]{
	\coordinate (start) at #1;
	\def \hs {#2}
	\def \ws {#3}
	\def \cs {#4}
	
	\begin{scope}[rotate around={90:(start)}]
		\draw[black] (start) -- ++(0,-\cs,0) -- ++(0,0,-\ws) -- ++(0,\cs,0);
		\draw[black] (start) -- ++(0,0,-\ws) -- ++(-\hs,0,0) node[left, pos=0.5]{\small #5} -- ++(0,0,\ws) node[left, pos=0.5]{\small #6} -- cycle;
		\draw[black] (start) -- ++(0,-\cs,0) -- ++(-\hs,0,0) -- ++(0,\cs,0) node[below, pos=0.5]{\small #7};
	\end{scope}
}

\newcommand{\conv}[7]{
	
}

\begin{tikzpicture}

	\def \lw {0.9mm}

    \coordinate (F1) at (1.4, 0, 0);
	\coordinate (F2) at (4.0, 0, 0);
	\coordinate (F3) at (8.2 ,0, 0);
	\coordinate (F4) at (10.9 ,0, 0);
	\coordinate (F5) at (13.5 ,0, 0);

	\node[inner sep=3pt] (cpn_f) at (22.5,-1.2) 
		{\includegraphics[scale=1.0]{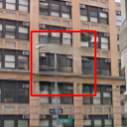}};

	\node[inner sep=3pt] (cpn_c) at (17.6,-1.2) 
		{\includegraphics[scale=1.0]{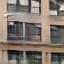}};

	\node[inner sep=3pt] (fpn_c) at (-6,-1.2) 
		{\includegraphics[scale=1.0]{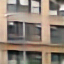}};

	\node[inner sep=3pt] (fpn_f) at (-11,-1.2) 
		{\includegraphics[scale=1.0]{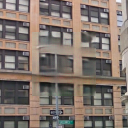}};

	\node[thin, outer sep=3pt](res) at (-1.8,-1.2)
			{\includegraphics[scale=1.0]{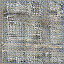}};

	\node[thin, circle, draw, outer sep=3pt](sum) at (-3.9,-1.2) {+};

	\draw[->, dashed,line width=\lw, black] (cpn_f) -- (cpn_c);
	\draw[->, dashed,line width=\lw, black] (fpn_c) -- (fpn_f);
	\draw[->, line width=\lw, black] (res) -- (sum);
	\draw[->, line width=\lw, black] (sum) -- (fpn_c);

	\path[->] (cpn_c) edge ++(-3,0,0);

	\node[] at (22.5,-3.9) {\huge Input};
	\node[] at (-1.7,-2.9) {\huge Residual};
	\node[] at (-1.7,-3.7) {\huge Image};
	\node[] at (-5.9,-2.9) {\huge Output};
	\node[] at (-11,-4.2) {\huge Inpainted};
	\node[] at (-11,-5) {\huge Image};

	\layer{(F5)}{3}{3}{0.7}{64}{64}{64}
	\draw[->, black] (F5) ++(-0.8,-1.0,0) -- ++(-1.0,0,0);
	\layer{(F4)}{3}{3}{0.7}{64}{64}{64}
	\draw[->, black] (F4) ++(-0.8,-1.0,0) -- ++(-1.0,0,0);
	\layer{(F3)}{3}{3}{0.7}{64}{64}{64}
	\draw[->, black] (F3) ++(-0.8,-1.0,0) -- ++(-1.0,0,0) node[left]{{\bf ...}}
		 ++(-0.5,0,0) -- ++(-1.0,0,0);
	\layer{(F2)}{3}{3}{0.7}{64}{64}{64}
	\draw[->, black] (F2) ++(-0.8,-1.0,0) -- ++(-1.0,0,0);
	\layer{(F1)}{3}{3}{0.7}{64}{64}{64}
	\draw[->, black] (F1) ++(-0.8,-1.0,0) -- ++(-1.0,0,0);

	\draw[->,line width=\lw, black] (cpn_c) -- ++(0,-5) -- ++(-21.5,0) -- (sum);



\end{tikzpicture}